\documentclass[Afour,times,sageh]{sagej}

\usepackage{moreverb,url}
\usepackage{amsmath}
\usepackage{subfig}
\usepackage{graphicx}
\usepackage{supertabular,enumitem}
\DeclareMathOperator*{\argmax}{argmax}

\usepackage[colorlinks,bookmarksopen,bookmarksnumbered,citecolor=red,urlcolor=red]{hyperref}

\newcommand\BibTeX{{\rmfamily B\kern-.05em \textsc{i\kern-.025em b}\kern-.08em
T\kern-.1667em\lower.7ex\hbox{E}\kern-.125emX}}

\def\volumeyear{2020}
\begin{document}

\runninghead{FordAV et al.}

\title{Ford Multi-AV Seasonal Dataset}

\author{Siddharth Agarwal*\affilnum{1}, Ankit Vora*\affilnum{1}, Gaurav Pandey\affilnum{2}, Wayne Williams\affilnum{1}, Helen Kourous\affilnum{1} and James McBride\affilnum{2}}

\affiliation{\affilnum{1}Ford AV LLC, Dearborn, MI, USA\\
\affilnum{2}Ford Motor Company, Dearborn, MI, USA}

\corrauth{*Equal Contributions - Siddharth Agarwal and Ankit Vora\\ Ford AV LLC,
20000 Rotunda Dr, Dearborn, MI 48124, USA}

\email{sagarw20@ford.com and avora3@ford.com}

\begin{abstract}
This paper presents a challenging multi-agent seasonal dataset collected by a fleet of Ford autonomous vehicles at different days and times during 2017-18. The vehicles traversed an average route of 66 km in Michigan that included a mix of driving scenarios such as the Detroit Airport, freeways, city-centers, university campus and suburban neighbourhoods, etc. Each vehicle used in this data collection is a Ford Fusion outfitted with an Applanix POS-LV GNSS system, four HDL-32E Velodyne 3D-lidar scanners, 6 Point Grey 1.3 MP Cameras arranged on the rooftop for 360-degree coverage and 1 Pointgrey 5 MP camera mounted behind the windshield for the forward field of view. We present the seasonal variation in weather, lighting, construction and traffic conditions experienced in dynamic urban environments. This dataset can help design robust algorithms for autonomous vehicles and multi-agent systems. Each log in the dataset is time-stamped and contains raw data from all the sensors, calibration values, pose trajectory, ground truth pose, and 3D maps. All data is available in Rosbag format that can be visualized, modified and applied using the open-source Robot Operating System (ROS). We also provide the output of state-of-the-art reflectivity-based localization for bench-marking purposes. The dataset can be freely downloaded at \href{avdata.ford.com}{avdata.ford.com}.
\end{abstract}


\maketitle

\section{Introduction}
In recent years, various companies and organizations have been working on one of the most challenging problem of transportation - a fully autonomous, self-driving vehicle. In order to navigate autonomously and make decisions, vehicles need to make sense of their environment. Many research groups have relied on camera based solutions on account of their high color resolution and inexpensive nature. On the other hand, 3D range sensors like light detection and ranging (LIDAR) scanners have become more appealing and feasible with the advances in consumer-grade technology. Furthermore, each sensor has its own pros and cons. For a vehicle to handle a variety of operating conditions, a more robust solution involving fusion of multiple sensors may be required. Systems today use a combination of 3D scanners, high resolution cameras and GPS/INS, to enable autonomy. No matter what sensors they use, autonomous vehicles will have to navigate in dynamic urban environments and successfully handle a variety of scenarios and operating conditions. These vehicles will have to negotiate with other autonomous and non-autonomous vehicles out on the road, thus opening up research avenues in Multi-Agent Autonomous Systems. This Multi-AV Seasonal dataset can provide a basis to enhance state-of-the-art robotics algorithms related to multi-agent autonomous systems and make them more robust to seasonal and urban variations. We hope that this dataset will be very useful to the Robotics and Artificial Intelligence community and will provide new research opportunities in collaborative autonomous driving.

\begin{figure}[ht]
	\centering
	\captionsetup{justification=centering}
	\subfloat{\includegraphics[width=0.75\linewidth]{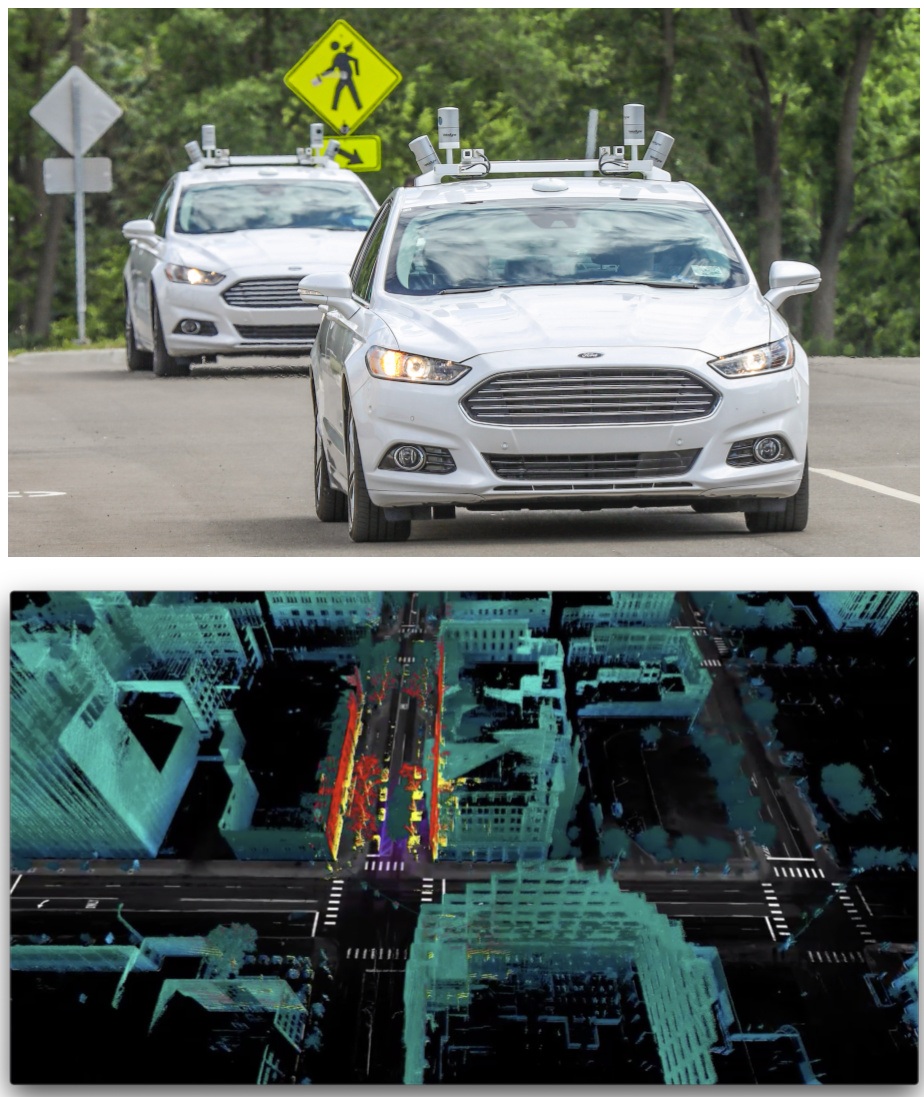}}
	\caption{The Ford Multi-AV Seasonal Dataset for long term autonomy} 
\end{figure}

\newpage

\section{Related Work} 
Within the past decade, there has been a significant advancement in autonomy. A lot of this development has been data centric for which the researchers rely on publicly available datasets. \cite{Pandey.McBride.Eustice2011} retrofitted a Ford F-250 pickup truck with HDL-64E lidar, Point grey Ladybug3 omnidirectional camera, Riegl LMS-Q120 lidar, Applanix POS LV and Xsens MTi-G IMU to release one of the first few publicly available datasets. This dataset includes small and large loop closure events, feature rich downtown and parking lots which makes it useful for computer vision and simultaneous localization and mapping algorithms. 

The KITTI dataset \citep{Geiger2013IJRR} is another benchmark dataset collected from an autonomous vehicle platform. The sensor data provided aims at progressive research in the field of optical flow, visual odometry, 3D object detection and tracking. Along with the raw data, the dataset also includes ground truth and benchmarking metrics \citep{KittiBenchmark} which enables evaluating a new algorithm against the state-of-the-art. The University of Oxford's robotcar dataset \citep{RobotCarDatasetIJRR} is another dataset rich in terms of variety in sensors and seasons. The focus of this dataset is to enable long term localization and mapping. The dataset contains data from stereo cameras, monocular cameras, lidars, GPS/IMU collected over a span of 17 months which includes all weather scenarios and construction scenarios. All of the above mentioned datasets do not provide any map information. The nuScenese dataset \citep{nuscenes2019} contains semantic maps which provide information on roads, sidewalks and crosswalks, to be used as a prior for object detection, tracking and localization. Argoverse \citep{Chang_2019_CVPR}, along with vehicle trajectories and 3D object bounding boxes, also contains maps of driveable area and vector maps of lane centerline and their connectivity. ApolloScape \citep{huang2018apolloscape}, CityScapes \citep{Cordts2016Cityscapes} and Mappilary \citep{MVD2017} are other datasets that focus on semantic segmentation using a combination of images and lidar. Such datasets also exists in other robotics domains like the CMU-YCB object and model set for benchmarking in robotic manipulation research \citep{sun2018pix3d}. One of the limitations of the datasets mentioned so far is that they are mostly collected from only one autonomous vehicle. 

Here we present a large scale multi-AV dataset augmented with 3D map of the environment. This would provide a significant database for autonomous vehicle research as well as multi-agent research including cooperative localization \citep{zhang2016cooperative}. 

The major contributions of this research are: 
\begin{enumerate}[noitemsep,partopsep=0pt,topsep=0pt,parsep=0pt]
\item[(i)]  A Multi-Agent dataset with seasonal variations in weather, lighting, construction and traffic conditions
\item[(ii)]  Full resolution time stamped data from 4 lidars and 7 cameras
\item[(iii)]  GPS data, IMU Pose trajectory and SLAM corrected ground truth pose
\item[(iv)]  High resolution 2D ground plane lidar reflectivity and 3D point cloud maps
\item[(v)]  State-of-the-art lidar reflectivity based localization \citep{Levinson.Thrun2007, Levinson10} results with ground truth for benchmarking
\item[(vi)]  All data can be visualized, modified and applied using the open-source Robot Operation System (ROS) \citep{ROS}
\end{enumerate}
The Ford Multi-AV Seasonal dataset consists of diverse scenarios that include:

\begin{enumerate}[noitemsep,partopsep=0pt,topsep=0pt,parsep=0pt]
\item[(i)] Seasonal/ Weather variation - sunny, cloudy, fall, snow

\item[(ii)] Traffic conditions - construction, oversized vehicles, pedestrians, congestion, under-pass, bridges, tunnels
\item[(iii)] Driving environments - highway, university campus, residential areas, airport pick-up/drop-off
\end{enumerate}

\section{Hardware Platform} 

We used a fleet of 2014 Ford Fusion Hybrids as the base platform. All the sensors were strategically placed on the vehicle as shown in \ref{fig:vehicle_sensor_config}. The trunk of the car was used to install four Quad -core i7 processors with 16 GB Ram, networking devices and a cooling mechanism. All the post processing was done on a Dell Precision 7710 laptop. The vehicle was integrated with the following sensors:

\begin{figure}
	\centering
	\captionsetup{justification=centering}
	\subfloat{\includegraphics[width=\linewidth]{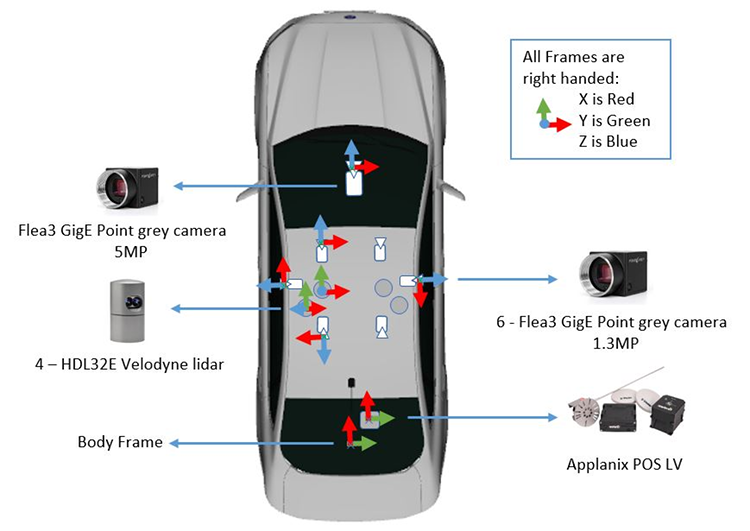}}\\
	\caption{Vehicle sensor configuration} 
	\label{fig:vehicle_sensor_config}
\end{figure}

\subsection{Velodyne HDL-32E Lidars} 

Each vehicle is equipped with a set of 4 Velodyne HDL-32E lidars \citep{velodyne} that are mounted on the roof. This lidar consist of 32 laser diodes providing an effective 40$^{\circ}$ vertical field of view. The entire unit can spin about its vertical axis to provide a full 360$^{\circ}$ azimuthal field of view. We captured our data set with the laser spinning at 10 Hz. The combined point cloud provides a 360 degree coverage around the car. We eliminate the points that hit the vehicle by designing a custom mask for each lidar.

\subsection{Flea3 GigE Point Grey Cameras}

Our fleet is equipped with 7 Flea3 GigE Point Grey Cameras \citep{camera_small}. These are CCD color cameras with 12 bit ADC and Global shutter. There are 2 stereo pairs of 1.3 MP cameras - Front Left-Right pair and a Rear Left-Right pair mounted on the roof of each vehicle. In addition to that, there are two 1.3 MP cameras on the sides and one 5 MP center dash camera mounted near the rear-view mirror. A combination of these 7 cameras provide an excellent field of view (80$^{\circ}$ for each 1.3 MP and 40$^{\circ}$ for the 5 MP). We captured images at 15 Hz for the 1.3 MP stereo pairs and side cameras, and 6 Hz for the center 5 MP camera. 

\subsection{Applanix POS LV}

Applanix \citep{applanix} is a professional-grade, compact, fully integrated, turnkey position and orientation system combining a differential GPS, an inertial measurement unit (IMU) rated with 1$^{\circ}$ of drift per hour, and a 1024-count wheel encoder to measure the relative position, orientation, velocity, angular rate and acceleration estimates of the vehicle. In our data set we provide the 6-DOF pose estimates obtained by integrating the acceleration and velocity.

\section{Sensor Calibration}

All the sensors on each vehicle are fixed and calibrated with respect to the origin of the vehicle i.e. the center of the rear axle. We provide intrinsic and extrinsic rigid-body transformation for each sensor with respect to this origin, also called as the body frame of the vehicle. The transformation is represented by the 6-DOF pose of a sensor coordinate frame where X\textsubscript{AB} = \([x, y, z, qx, qy, qz, qw]\) \citep{Smith1988} denotes the 6-DOF pose of frame A (child) to frame B (parent). We follow the same procedure for each respective sensor on all the vehicles in the fleet. The calibration procedures are summarized below. 

\subsection{Applanix to Body Frame}

The Applanix is installed close to the origin of the body frame, which is defined as the center of the rear axle of the vehicle. A research grade coordinate measuring machine (CMM) was used to precisely obtain the position of the applanix with respect to the body frame. Typical
precision of a CMM is of the order of micrometers, thus for all
practical purposes we assumed that the relative position obtained from CMM are true values without any error. We provide the 6 DOF position and orientation of the applanix relative to the body frame for each vehicle given by X\textsubscript{ab} = \([x, y, z, qx, qy, qz, qw]\).

\subsection{Lidar Calibration}
\subsubsection{Intrinsics}

The dataset includes reflectivity calibration information for each of the HDL-32E lidars for each vehicle \citep{levinson2014unsupervised} and beam offsets as provided by Velodyne \citep{velodyne}. Each laser beam of the lidar has variable reflectivity mapping. As a result, we also provide reflectivity calibration files for each lidar.

\subsubsection{Extrinsics}

To start with an estimate or lidar positions in 6DOF, we used CMM to precisely obtain the position of some known
reference points on the car with respect to the body frame,
The measured CMM points are denoted X\textsubscript{bp}. We then manually measured the position of each lidar from one of these reference points to get X\textsubscript{pl}. The relative transformation of the lidar with
respect to the body frame is thus obtained by compounding
the two transformations \citep{Smith1988}. This estimate is used to initialize the Generalized Iterative Closest Point (ICP) Algorithm \citep{segal2009generalized}, which matches the lidar scans from each lidar against other lidars to obtain a centimeter level accurate transformation from lidar to body given by X\textsubscript{lb} = \([x, y, z, qx, qy, qz, qw]\) for each lidar.

\subsection{Camera Calibration}
\subsubsection{Intrinsic}

calibration was performed on each of the cameras using the method described in AprilCal \citep{richardson2013}. This is an interactive suggestion based calibrator that performs real-time detection on feducial markers. All images in the dataset are undistorted using the camera intrinsic matrix and distortion coefficients as shown below:

{\centering
\begin{math}
K =
\begin{bmatrix}
f\textsubscript{x} & 0 & x\textsubscript{0} \\
0 & f\textsubscript{y} & y\textsubscript{0} \\
0 & 0 & 1
\end{bmatrix}
\hspace{15pt}
D = 
\begin{bmatrix}
D\textsubscript{n} & D\textsubscript{n-1} & ... & D\textsubscript{0}
\end{bmatrix}
\end{math}
\par }

\vspace{2pt}

where f\textsubscript{x} and f\textsubscript{y} are the focal lengths, x0 and y0 are the principal point offsets and D is the set of distortion coefficients . We provide ROS format yaml files with the camera \((K)\), rotation \((R)\) and projection \((P)\) matrices. 
\subsubsection{Extrinsic}

calibration was performed to find the relative position of each camera with respect to the body frame of car. We use the method described in \cite{Pandey2012}. This is a mutual information based algorithm that provides a relative transformation from the camera to the lidar (X\textsubscript{cl}). We use lidar extrinsics (X\textsubscript{lb}) to finally compute the position of the camera relative to the body frame (X\textsubscript{cb}).
\vspace{1em}

{\centering 
\begin{math}
X\textsubscript{cb} = X\textsubscript{lb} \bigoplus X\textsubscript{cl} 
\end{math}
\par }

\section{Data}

\begin{figure}[h]
	\centering
	\captionsetup{justification=centering}
	\subfloat{\includegraphics[width=0.81\linewidth]{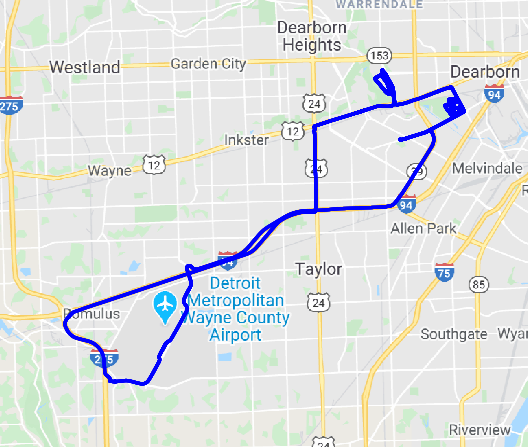}}\\
	\caption{Sample Trajectory}
	\label{fig:sampleTrajectory}
\end{figure}

All data was collected by a fleet of Ford fusion vehicles that were outfitted with a Applanix POS-LV inertial measurement unit (IMU), four HDL-32 Velodyne 3D-lidar scanners, 2 Point Grey 1.3 MP stereo camera pairs, 2 Point Grey 1.3 MP side cameras and 1 Point Grey 5 MP dash camera. The vehicles traversed an average route of 66 km in Michigan that included a mix of driving scenarios such as the Detroit Airport, freeways, city-centers, university campus and suburban neighbourhoods, etc. A sample trajectory of one vehicle in one of the runs is shown in \ref{fig:sampleTrajectory}.

This multi-agent autonomous vehicle data presents the seasonal variation in weather, lighting, construction and traffic conditions experienced in dynamic urban environments. The dataset can help design robust algorithms for autonomous vehicles and multi-agent systems. Each log in the dataset is time-stamped and contains raw data from all the sensors, calibration values, pose trajectory, ground truth pose and 3D Maps. All data is available in Rosbag format that can be visualized, modified and applied using ROS.  We also provide the output of state-of-the-art reflectivity-based localization \citep{Levinson.Thrun2007, Levinson10} with cm level accuracy for benchmarking purposes. Each parent folder in the Data Directory represents a drive that is marked by the date. Each drive has sub-directories corresponding to the vehicles used and the corresponding 3D Maps. Each vehicle sub-directory contains a Rosbag with all the data associated with that vehicle in that drive. We also provide scripts and commands to convert all this data to human readable format. The dataset is freely available to download at \href{avdata.ford.com}{avdata.ford.com}, a collaboration with the AWS Public Dataset Program \citep{aws}. 

Each rosbag contains the standard ROS messages as described in Table \ref{T1}

\begin{table*}[ht]
\small\sf\centering
\caption{Summary of messages in each Rosbag.\label{T1}}
\begin{tabular}{lllll}
\toprule
Source & Type & ROS Message & ROS Topic & Max Frequency (Hz)\\
\midrule
Red Lidar & 3D Scan & velodyne\textunderscore msgs/VelodyneScan & /lidar\textunderscore red\textunderscore scan & 10\\
Green Lidar & 3D Scan & velodyne\textunderscore msgs/VelodyneScan & /lidar\textunderscore green\textunderscore scan & 10\\
Blue Lidar & 3D Scan & velodyne\textunderscore msgs/VelodyneScan & /lidar\textunderscore blue\textunderscore scan & 10\\
Yellow Lidar & 3D Scan & velodyne\textunderscore msgs/VelodyneScan & /lidar\textunderscore yellow\textunderscore scan & 10\\
Front-Left Camera & 1.3 MP Image Thumbnail & sensor\textunderscore msgs/Image & /image\textunderscore front\textunderscore left & 15\\
IMU & IMU & sensor\textunderscore msgs/Imu & /imu & 200\\
GPS & 3D GPS & sensor\textunderscore msgs/NavSatFix & /gps & 200\\
GPS Time & GPS Time & sensor\textunderscore msgs/TimeReference & /gps\textunderscore time & 200\\
Raw Pose & 3D Pose & geometry\textunderscore msgs/PoseStamped & /pose\textunderscore raw & 200\\
Localized Pose & 3D Pose & geometry\textunderscore msgs/PoseStamped & /pose\textunderscore localized & 20\\
Ground Truth Pose & 3D Pose & geometry\textunderscore msgs/PoseStamped & /pose\textunderscore ground\textunderscore truth & 200\\
Raw Linear Velocity & 3D Velocity & geometry\textunderscore msgs/Vector3Stamped & /velocity\textunderscore raw & 200\\
Transforms & Pose Transformations & tf2\textunderscore msgs/TFMessage & /tf & 200\\
\bottomrule
\end{tabular}\\[10pt]
\end{table*}

\subsection{3D Lidars Scans}

Each vehicle is equipped with 4 HDL-32E lidars and their data is provided as standard VelodyneScan ROS messages. We designate these lidars as - Yellow, Red, Blue and Green going left to right when seen from the top.

\subsection{Camera Images}

Each rosbag contains images from all 7 cameras from the vehicle. The two front and the two rear cameras are 1.3 MP stereo pairs operating at a maximum rate of 15 Hz. The two 1.3 MP side cameras also work at 15 Hz. The front dash camera produces 5 MP images at a maximum rate of 6 Hz. All images have been rectified using the intrinsic parameters and stored as png images. All 1.3 MP cameras are triggered at the same time such that there is a image corresponding to each camera with the same timestamp. This dataset used an automated tool \citep{anonymizer} to blur vehicle licence plates and people's faces from all camera images. 

\subsection{IMU}

The IMU data consists of linear acceleration and angular velocity in \(m/s^2\) and \(rad/s\). These values represent the rate of change of the body frame. The IMU frame is oriented exactly like the body frame.

\subsection{GPS}

The GPS data provides the latitude, longitude and altitude of the body frame with respect to the WGS84 frame. The data is published at 200 Hz but can be sub-sampled to simulate a cheaper GPS. We also provide the GPS time as seconds from top of the week. This helps in syncing logs from multiple vehicles driven at the same time.

\subsection{Global 3D Maps} \label{globalMap}

\begin{figure*}[t]

\centering
\captionsetup{justification=centering}
\subfloat[]{\includegraphics[height=3cm,width=0.3\textwidth]{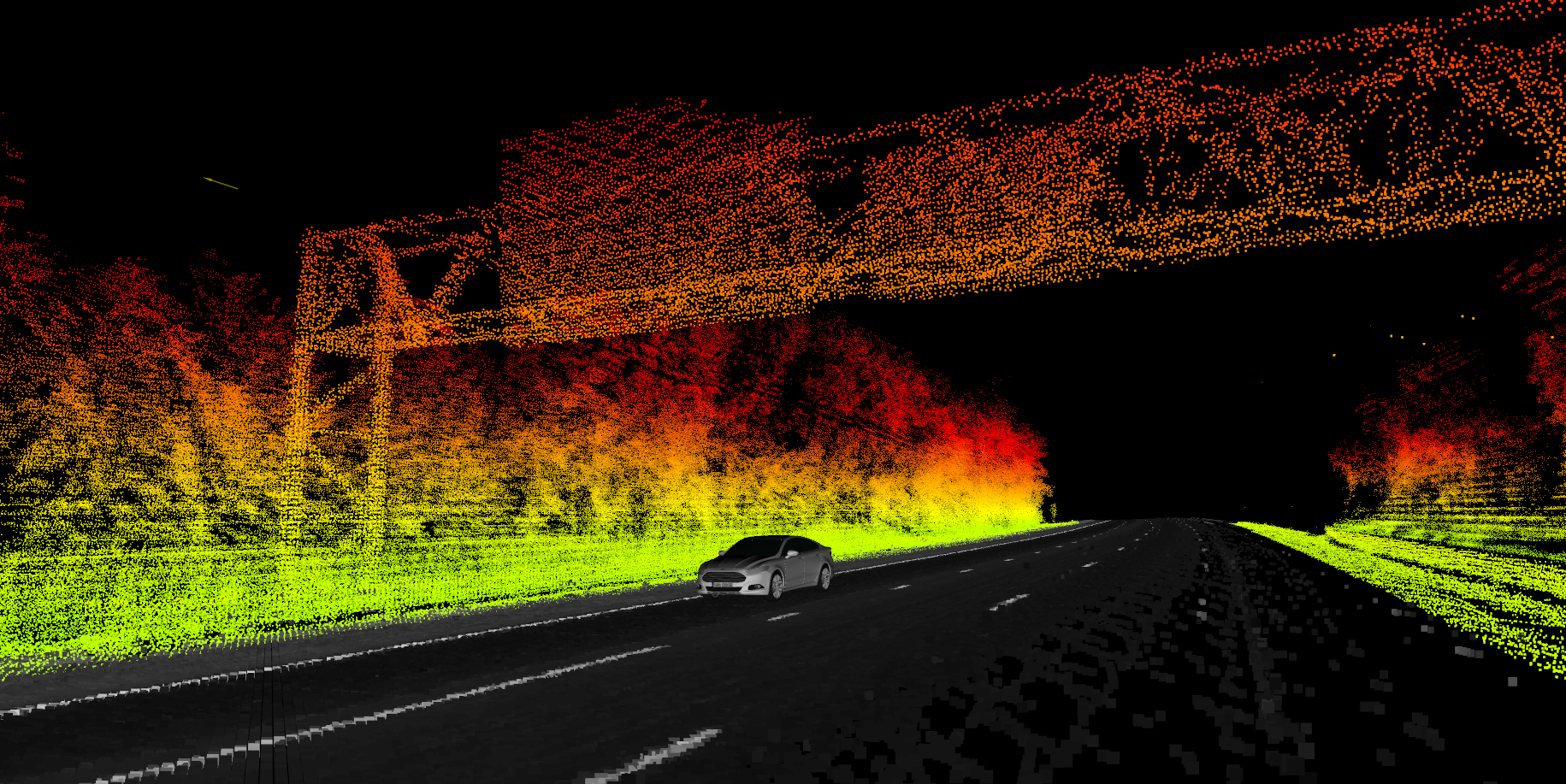}}
\hspace{1.5em}
\subfloat[]{\includegraphics[height=3cm,width=0.3\textwidth]{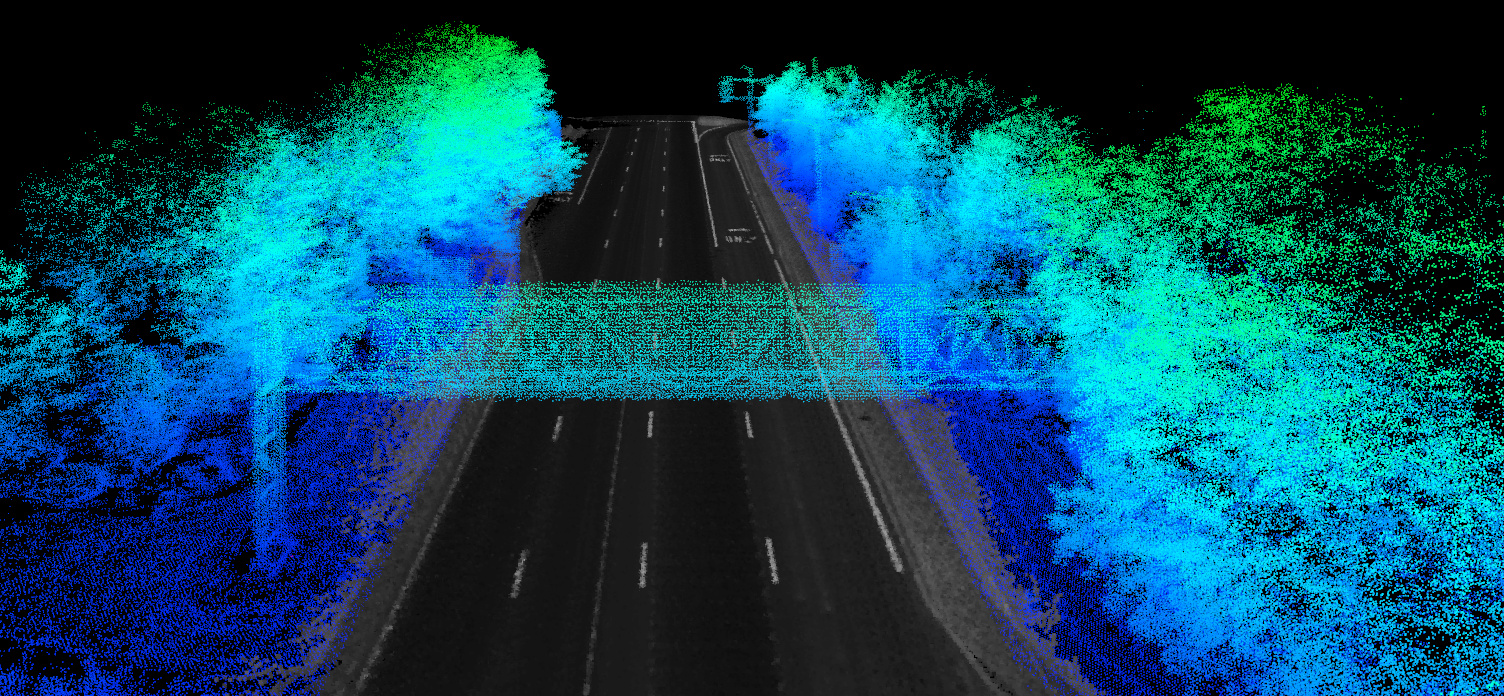}}
\hspace{1.5em}
\subfloat[]{\includegraphics[height=3cm,width=0.3\textwidth]{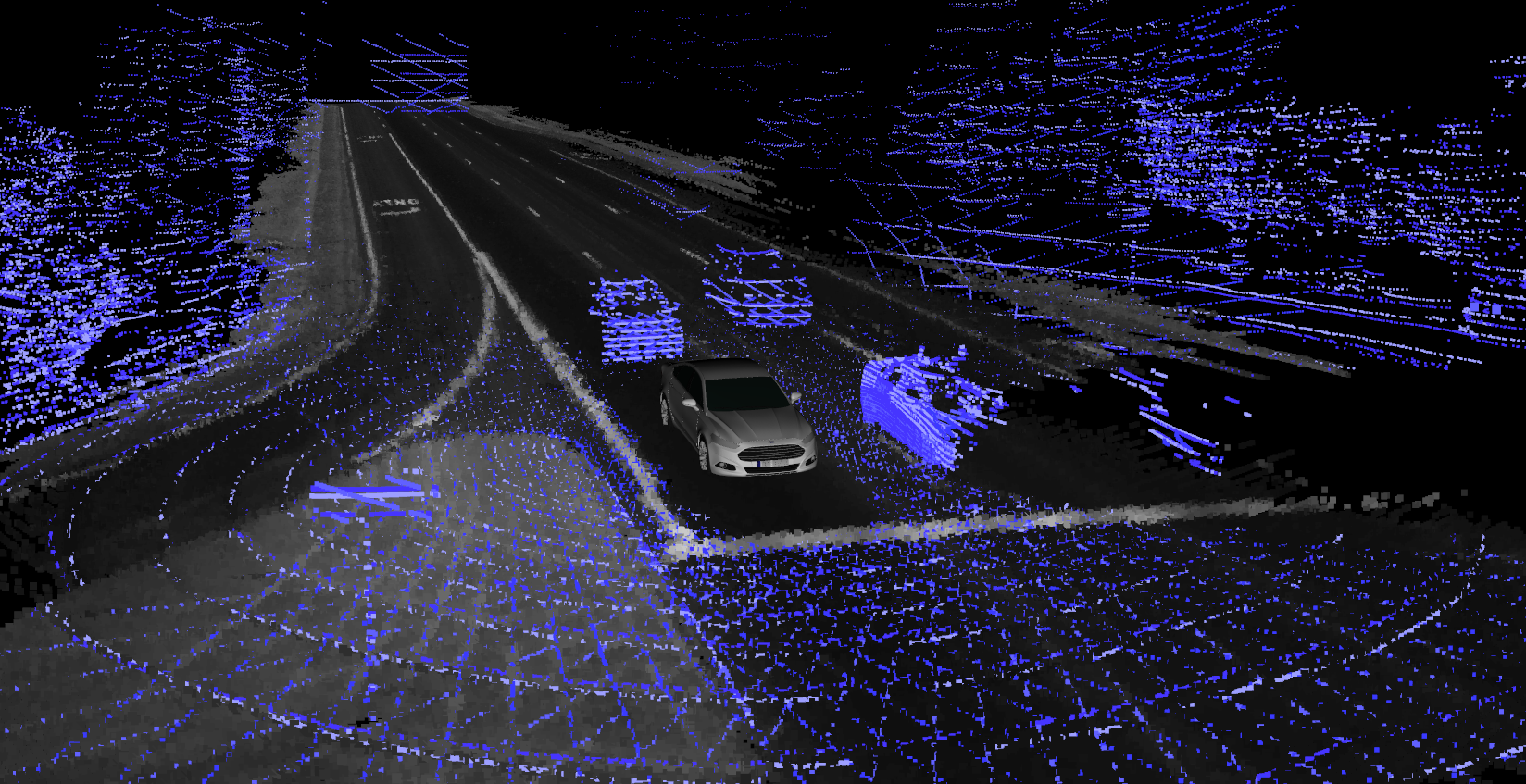}}

\caption{Sample output from visualization and development tools. (a) RViz visualization of the map and sensor data. (b) Reflectivity map and 3D pointcloud map. (c) Multi lidar live data visualization}
\label{fig:softwareTools}

\end{figure*}

Each drive in the dataset is accompanied by two types of global 3D maps - ground plane reflectivity and a map of 3D point cloud of non-road points as shown in Figure \ref{fig:softwareTools}. These maps are provided in open source PCD format \citep{pcl}. Estimation of the global prior-map modeling the environment uses the standard maximum a posteriori (MAP) estimate given the position measurements from odometry and various sensors \citep{Durrant-Whyte.Bailey2006, Bailey.Durrant-Whyte2006}. We use pose-graph SLAM with lidars to produce a 3D map of the environment. A pose-graph is created with odometry, 3D lidar scan matching and GPS constraints. Here, the lidar constraints are imposed via generalized iterative closest point (GICP)\citep{segal2009generalized}. To minimize the least squares problem we use the implementation of incremental smoothing and mapping (iSAM)\citep{Kaess.Ranganathan.Dellaert2008} in an offline fashion. From the optimized pose-graph, we create a dense ground plane and a full 3D point cloud which includes buildings, vegetation, road signs etc. by accumulating points in corresponding cells of a 3D grid. Since the \(3\sigma\) range of the lidar reflectivity is between 0 and 100, we scale the values linearly to 0-255 to cover the range, where 0 represents no data. This is also reflected in the the local relfectivity map. 

\subsection{Localization}

Our localization framework is based on using measurements from 3D lidar scanners to create a 2D grid local map of reflectivity around the vehicle. Localization is run online by matching the local grid map with the prior map along the x-y plane, also called image registration. The cell values within this local grid which we chose to be of size $40 \times 40$ m. (with a 10 cm cell resolution) represent the property of the world same as that stored in global maps. The local grid map is computed and updated online from the accumulated lidar points which are motion compensated using inertial measurements. 

\begin{figure}[ht]
	\centering
	\captionsetup{justification=centering}
	\subfloat[Lateral error]{\includegraphics[width=0.81\linewidth]{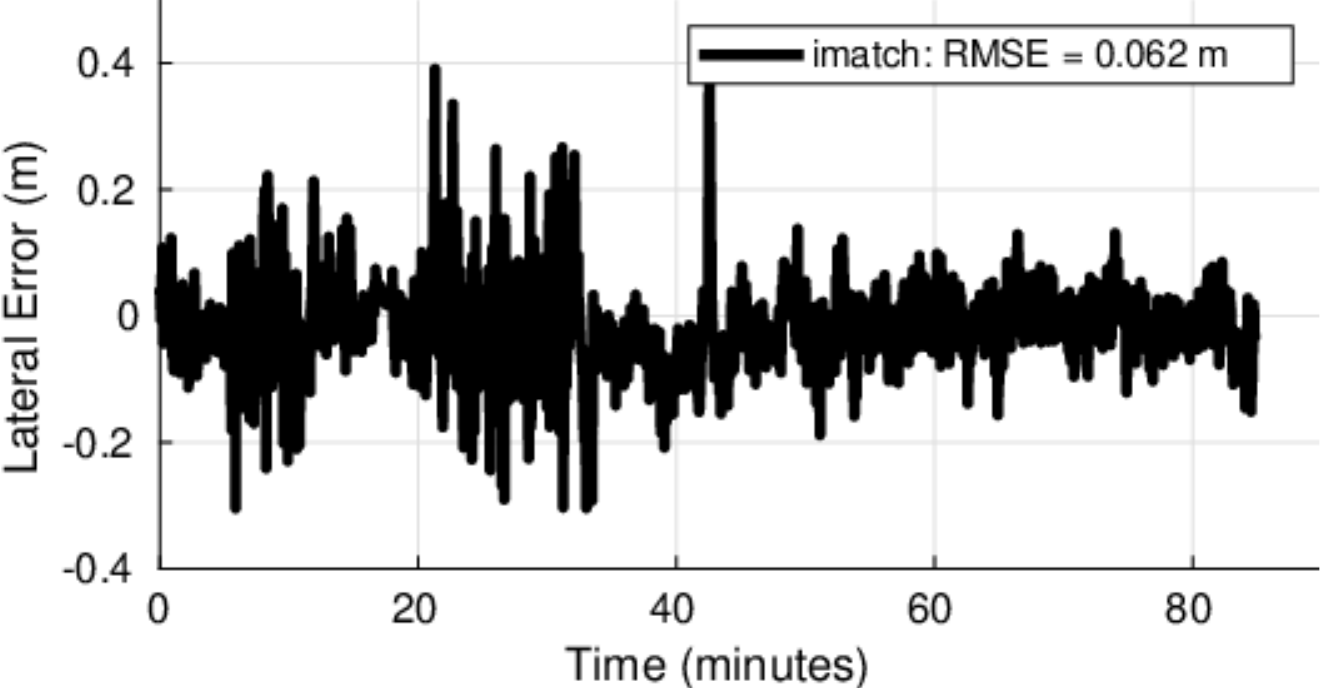}}
	
	\subfloat[Longitudinal error]{\includegraphics[width=0.81\linewidth]{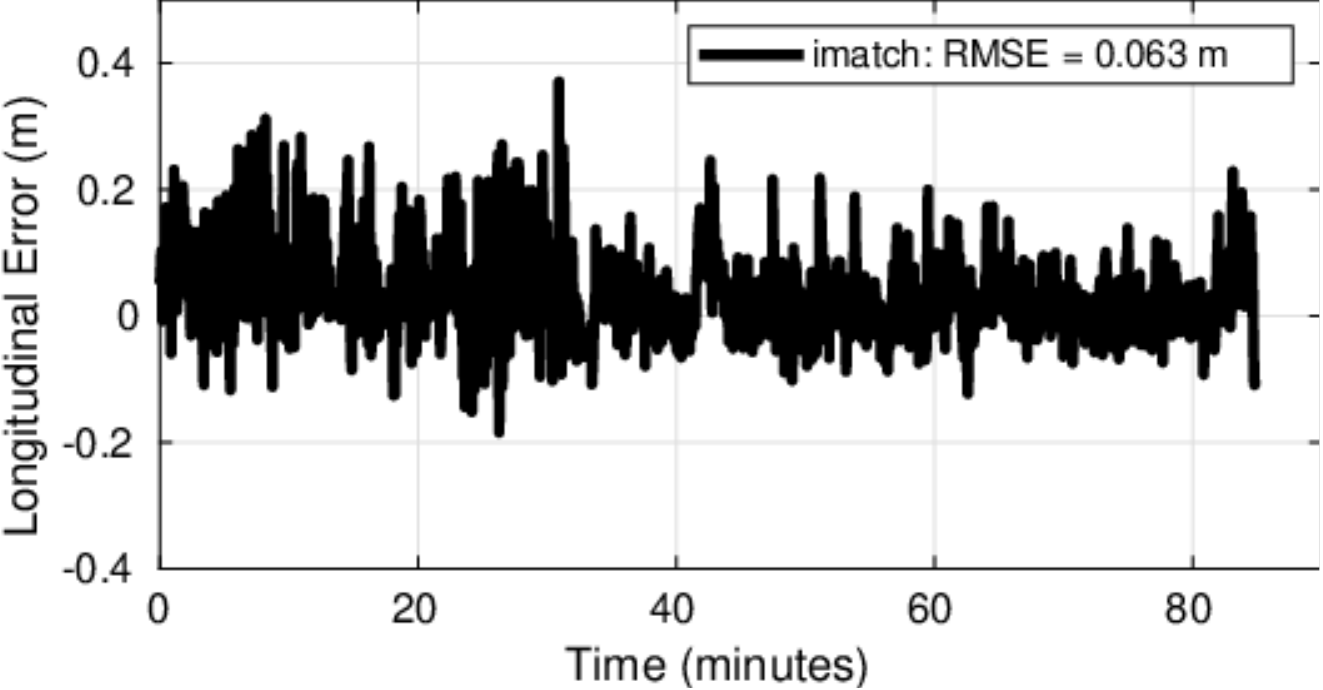}}
	
	\subfloat[Heading error]{\includegraphics[width=0.81\linewidth]{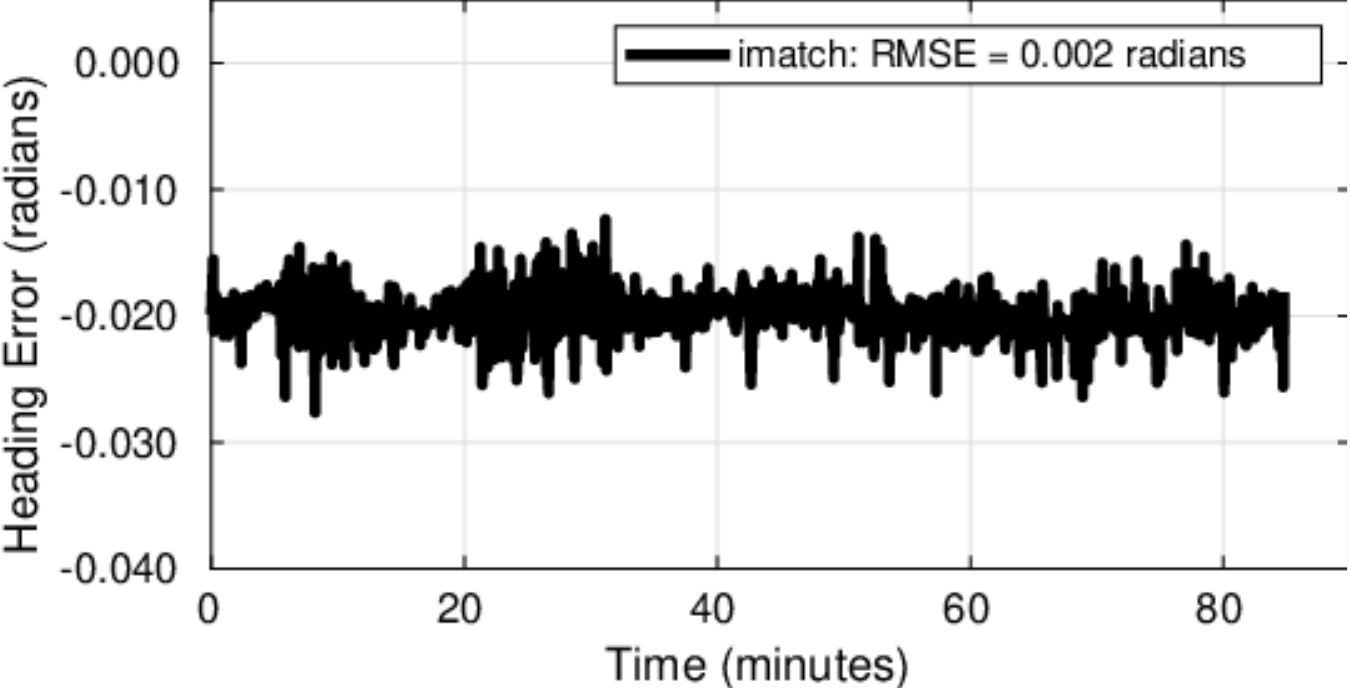}}
	\caption{Localization errors}
	\label{fig:performance}
\end{figure}

The planar GPS constraint of zero height in the optimization of the global prior-map described in section \ref{globalMap} simplifies the registration problem to a 3-DOF search over the \(x\), \(y\) and \(yaw (h)\) vehicle pose. Here, we propose to maximize the normalized mutual information (NMI)\citep{Studholme1999} between reflectivity of candidate patches from the prior map \(M\) and the local grid-map \(L_i\) \citep{wolcott2014visual}: \[(\hat{x},\hat{y},\hat{h}) = \argmax_{x \in \mathcal{X},y \in \mathcal{Y},h \in \mathcal{H}}NMI (M, L_i),\] where \(\mathcal{X}\), \(\mathcal{Y}\) and \(\mathcal{H}\) represent our search space in all 3 degrees of freedom.
The map matching is agnostic to the filter we use to fuse data from various sensors. We use an EKF filter to mainly localize the vehicle in 3 DOF with state vector $\mu_t$ = \(\{x_t, y_t, h_t\}\). In addition, the correction in \(z\) direction is accomplished using prior map lookup after the image registration corrections are obtained in \(x\) and \(y\). We use the standard EKF Predict and Update equations where \(F_t\) represents the state transition matrix obtained from Applanix GNSS solution, \(H_t\) represents the linearized measurement model, \(z_t\) and \(R_t\) represent the measurement and the uncertainty obtained from the image registration process respectively and \(K_t\) represents the Kalman gain.
We show the results of this localization technique for a sample log measured against the ground truth in Figure \ref{fig:performance}. The localization error is well within the bounds of requirements for autonomous vehicles as defined by \cite{Reid_2019}

\subsection{Ground Truth Pose}

An important contribution of this work is that we provide ground truth pose for all our logs as standard ROS Pose messages. This can help researchers design their own localization algorithm and calculate the 6 Degrees of Freedom errors against the ground truth. The ground truth pose is generated using full bundle adjustment; this method has been popularly used in the community \citep{  castorena2017ground, Wolcott17}. 

\section{Software and Tools}

For easy visualization and development, we provide a set of ROS packages on our \href{https://github.com/Ford/AVData}{Github} repository. These have been tested on Ubuntu 16.04, 32GB RAM and ROS Kinetic Kame. The sample output of the easy to use tools is shown in Figure \ref{fig:softwareTools}. We also provide scripts and commands to convert all the data to human readable format. 

\subsection{ford\textunderscore demo}

This package contains the roslaunch files, rviz plugins and extrinsic calibration scripts. The \textit{demo.launch} file loads the maps, rviz, and the vehicle model. In addition, it also loads the sensor to body extrinsic calibration files from the specified folder.
\textbf{Usage}: \textit{roslaunch ford\textunderscore demo demo.launch map\textunderscore dir:=/path/to/your/map calibration\textunderscore dir:=/path/to/your/calibration/folder/}

\subsection{fusion\textunderscore description}

This package contains the Ford fusion URDF for visualization in Rviz. The physical parameters mentioned in the URDF are just for representation and visualization and do not represent the actual properties of a Ford Fusion vehicle.

\subsection{map\textunderscore loader}

Map loader package loads the ground plane reflectivity and 3D pointcloud maps as ROS PointCloud2 messages. The package subscribes to vehicle pose to decide what section of the map to display. Various dynamic parameters include \textit{publish\textunderscore rate, pcd\textunderscore topic, pose\textunderscore topic, neighbor\textunderscore dist}. By default, the reflectivity map publishes on the \textit{/reflectivity\textunderscore map} topic and the 3D pointcloud is published on the \textit{/pointcloud\textunderscore map} topic. \textbf{Usage}: \textit{roslaunch map\textunderscore loader reflectivity\textunderscore map\textunderscore demo.launch map\textunderscore folder:=/path/to/your/map}

\begin{figure}
	\centering
	\captionsetup{justification=centering}
	\subfloat[Summer]{\includegraphics[width=0.45\linewidth, trim=0 0 0 60, clip]{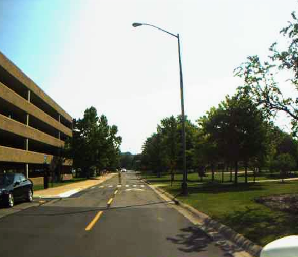}}
	\hspace{0.6em}
	\subfloat[Fall]{\includegraphics[width=0.45\linewidth, trim=0 0 0 60, clip]{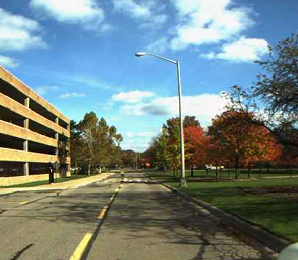}}

	\subfloat[Snow]{\includegraphics[width=0.45\linewidth, trim=0 0 0 60, clip]{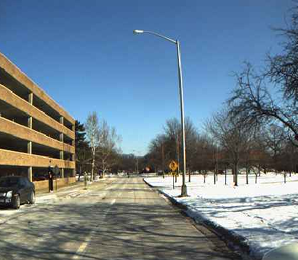}}
    \hspace{0.6em}
    \subfloat[Cloudy]{\includegraphics[width=0.45\linewidth, trim=0 0 0 90, clip]{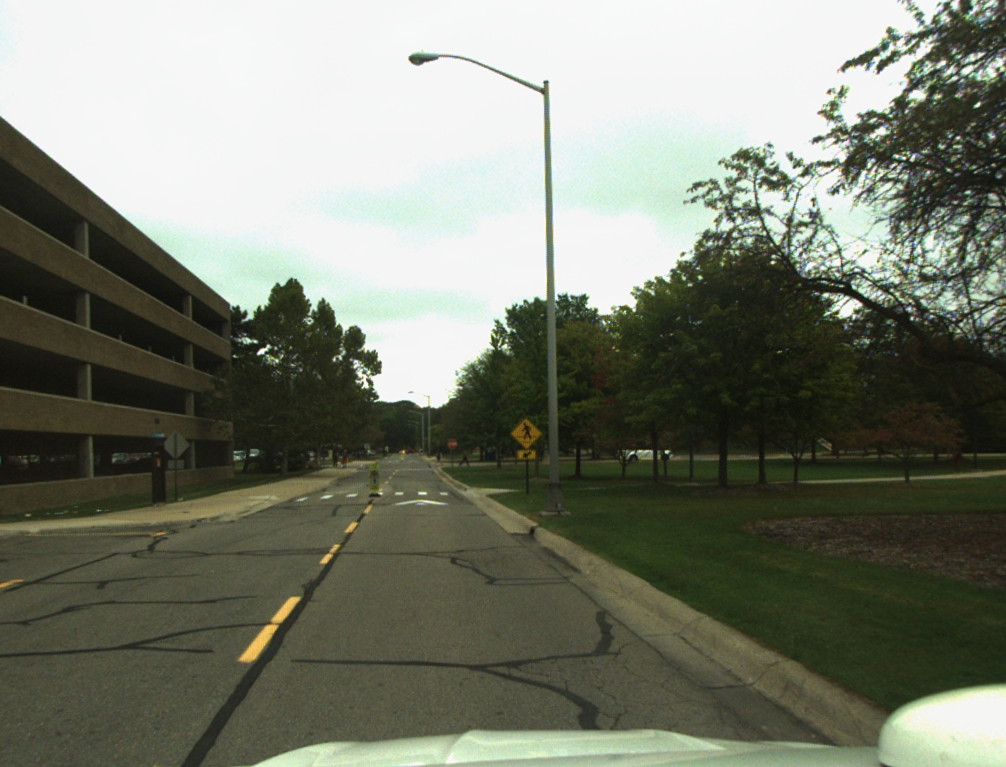}}
	\caption{Weather Variation}
	\label{fig:weather_variation}
\end{figure}

\begin{figure}
	\centering
	\captionsetup{justification=centering}
	\subfloat[Interstate]{\includegraphics[width=0.45\linewidth]{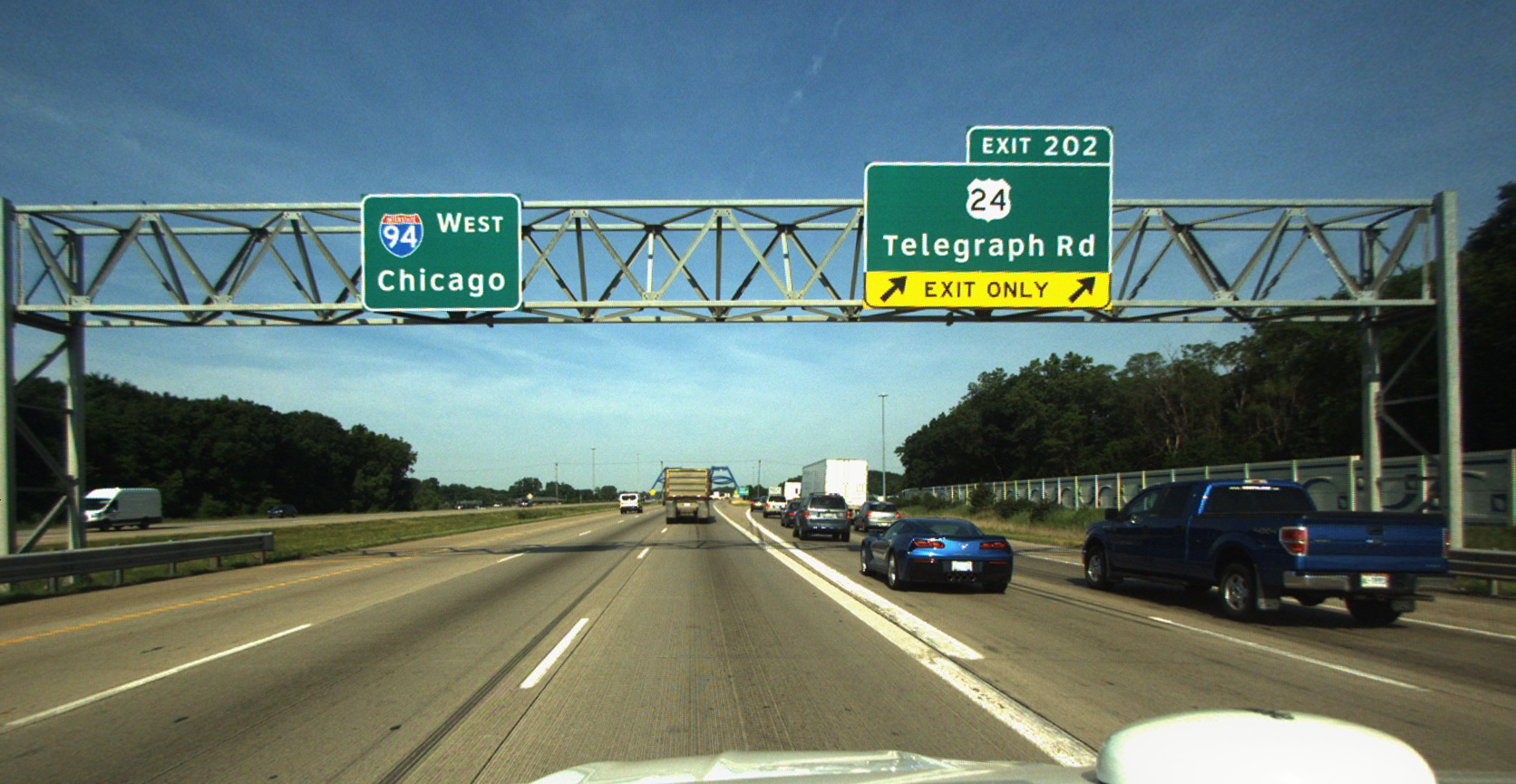}}
	\hspace{0.6em}
	\subfloat[Residential]{\includegraphics[width=0.45\linewidth]{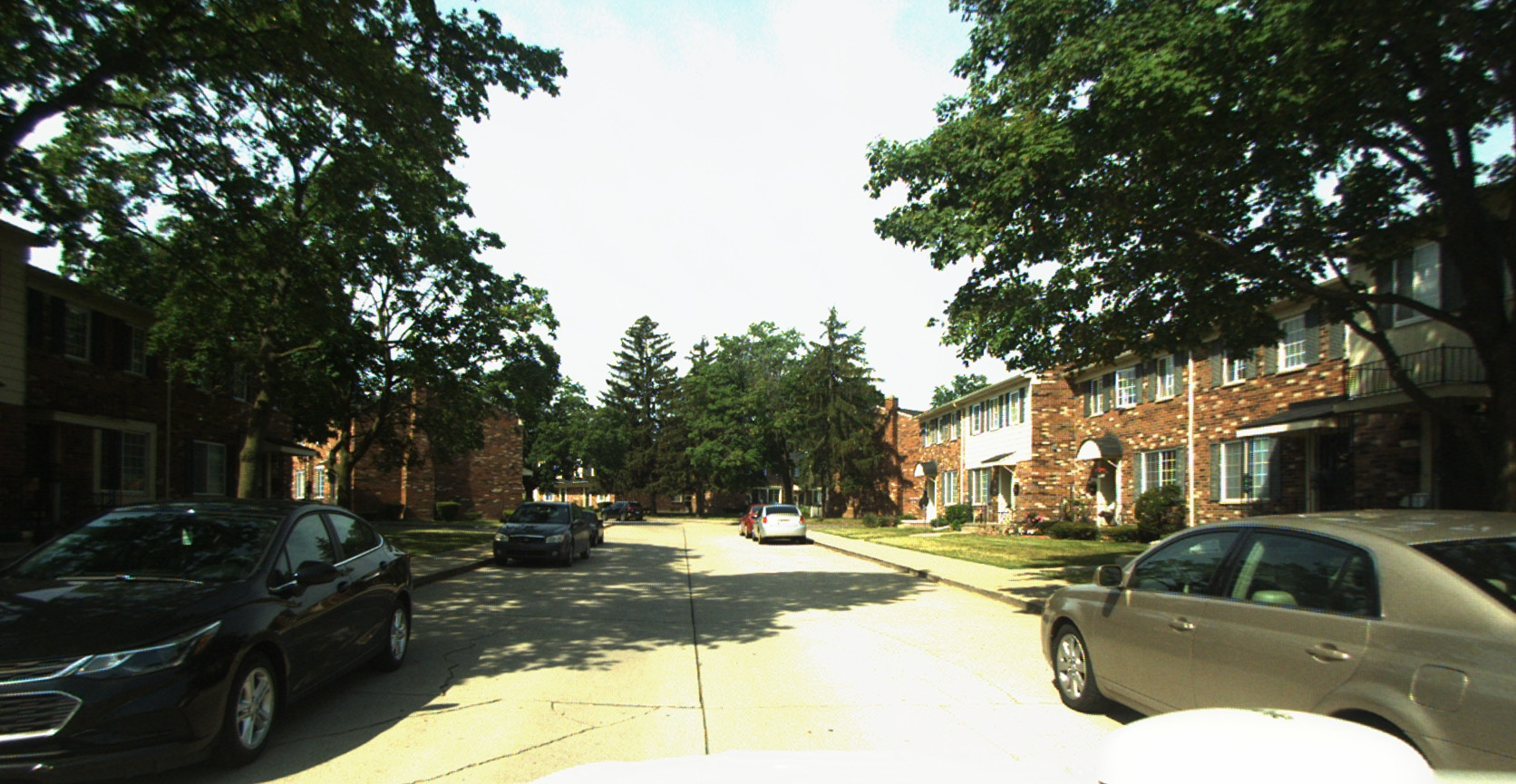}}

	\subfloat[Overpass]{\includegraphics[width=0.45\linewidth]{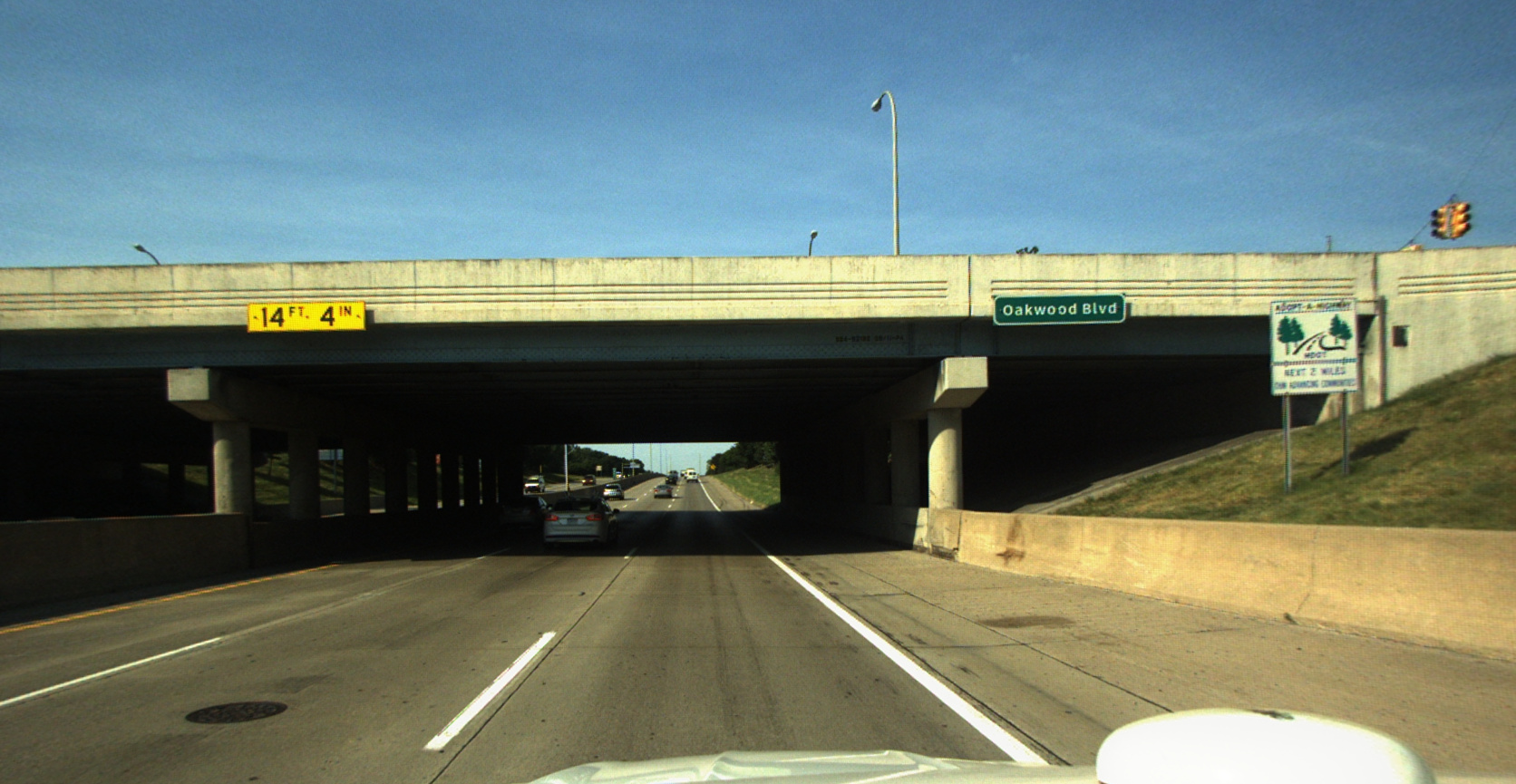}}
    \hspace{0.6em}
    \subfloat[Airport]{\includegraphics[width=0.45\linewidth]{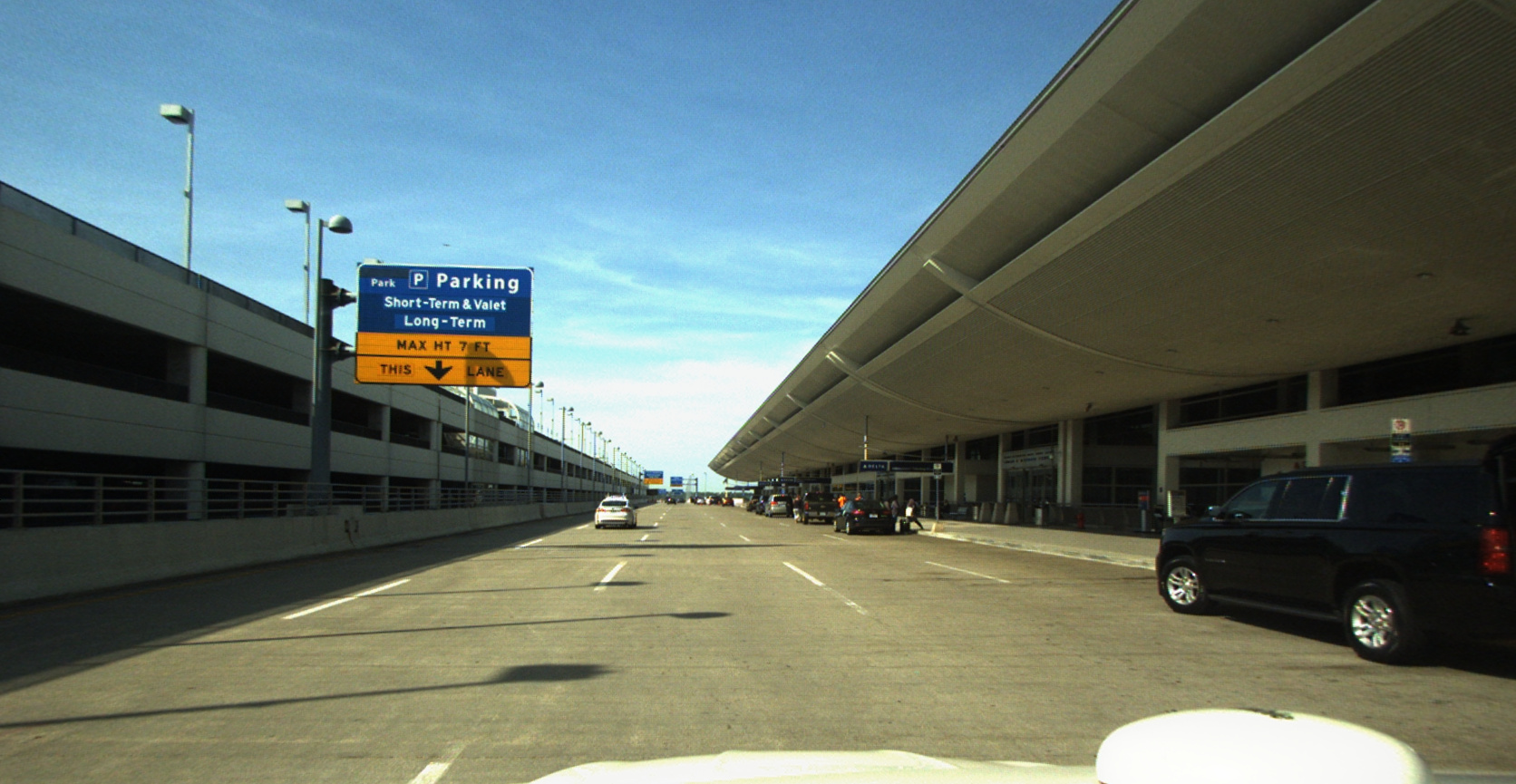}}

	\subfloat[Bridge]{\includegraphics[width=0.45\linewidth]{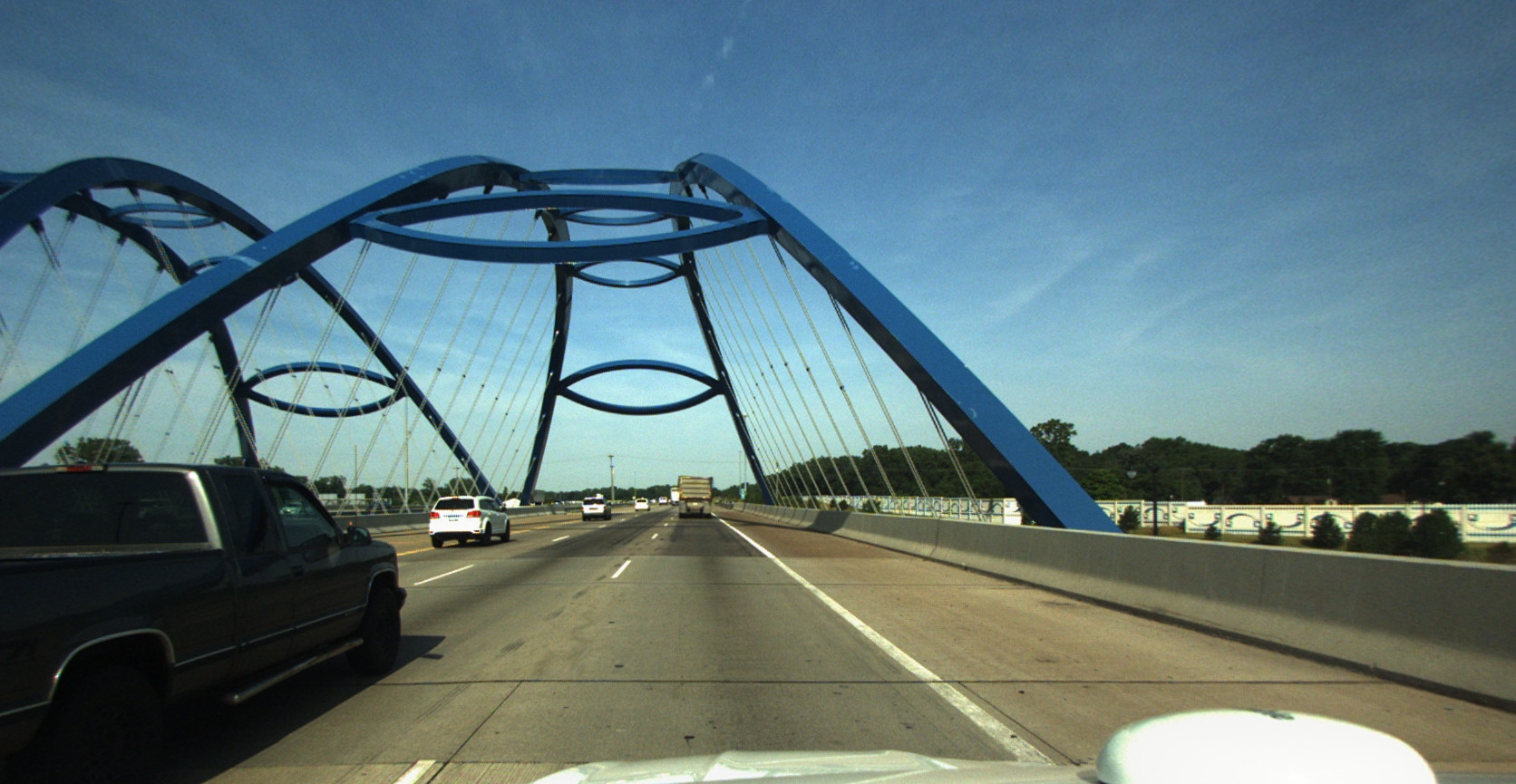}}
    \hspace{0.6em}
    \subfloat[Tunnel]{\includegraphics[width=0.45\linewidth]{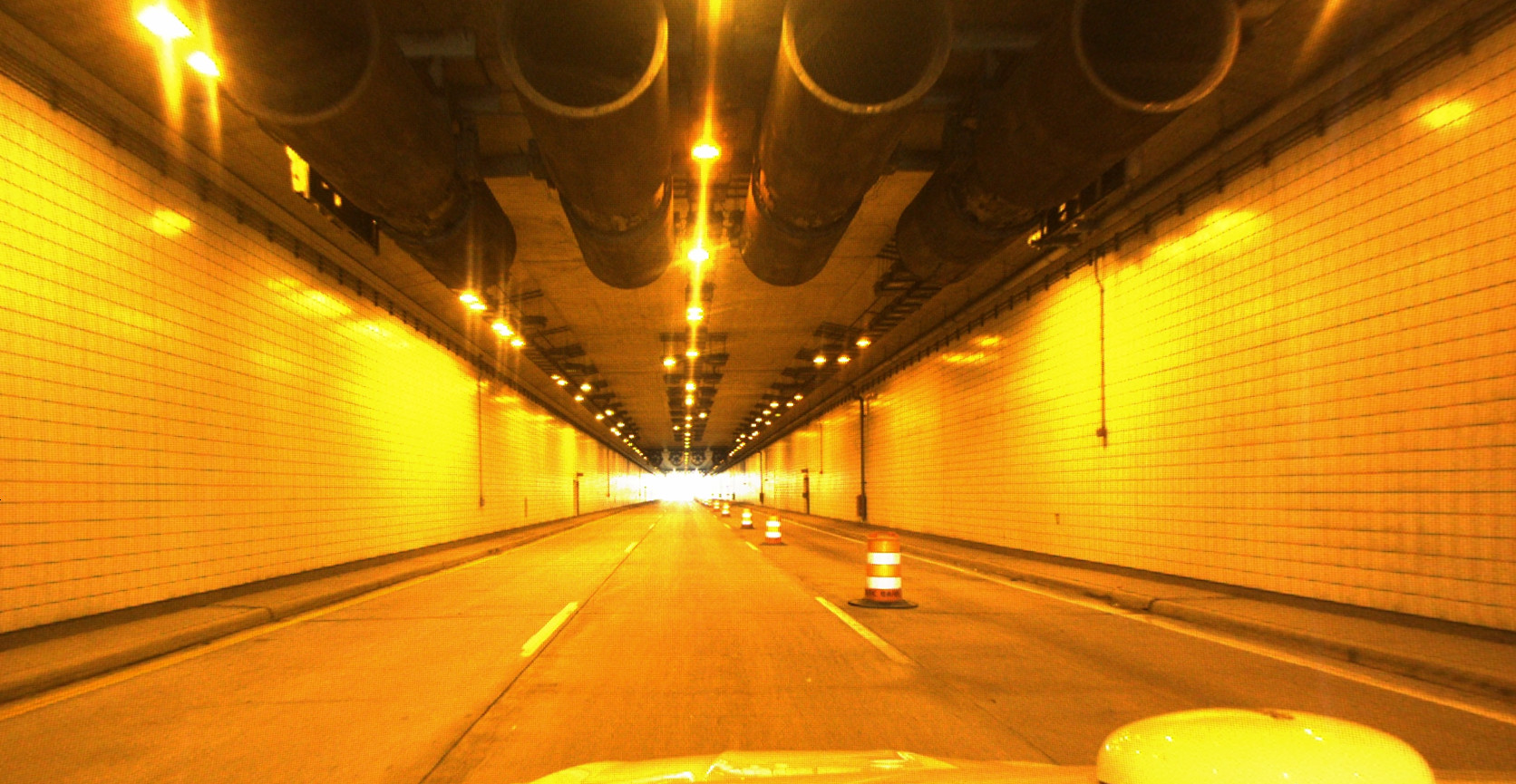}}
    
    \subfloat[Construction]{\includegraphics[width=0.45\linewidth]{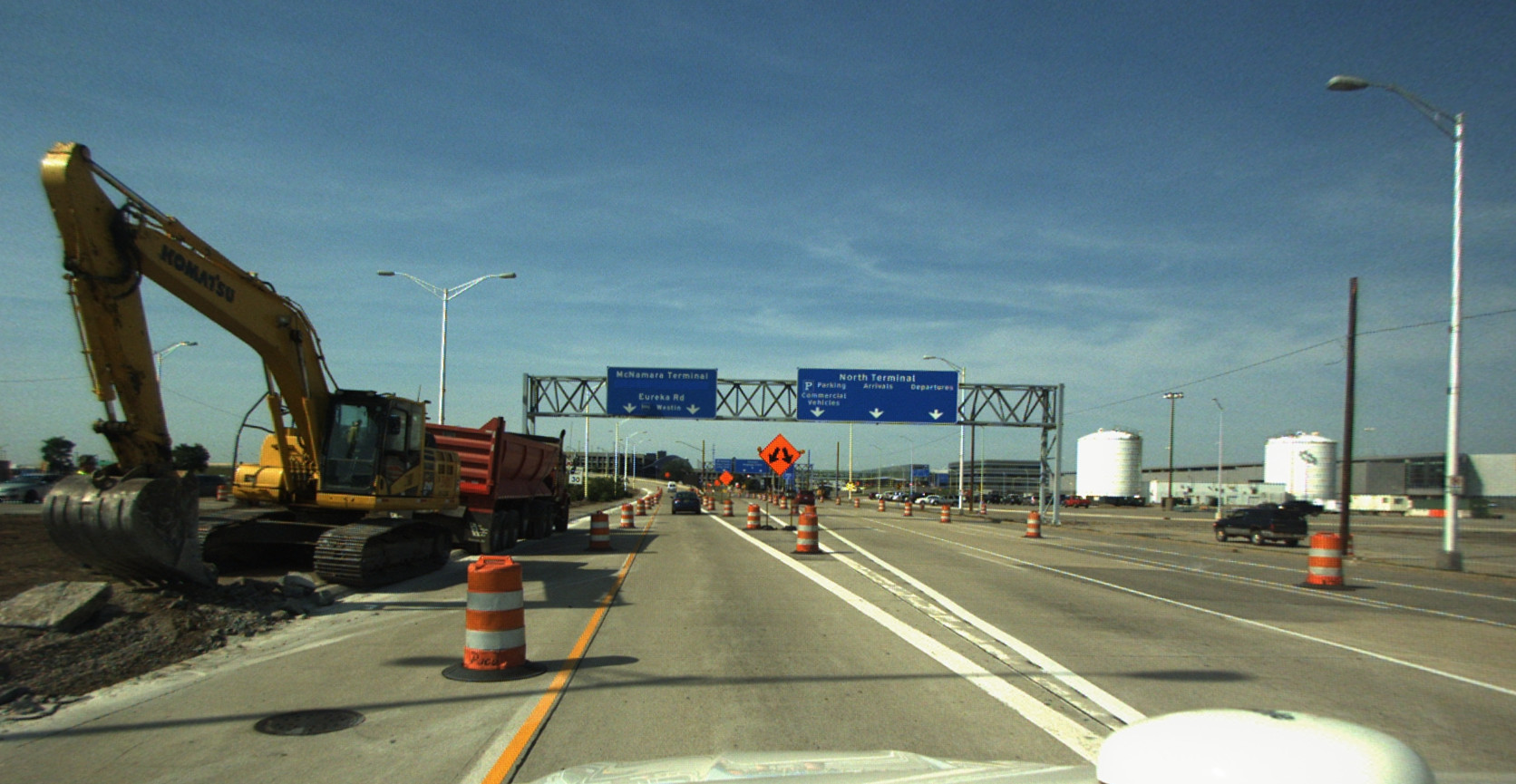}}
    \hspace{0.6em}
    \subfloat[Vegetation]{\includegraphics[width=0.45\linewidth]{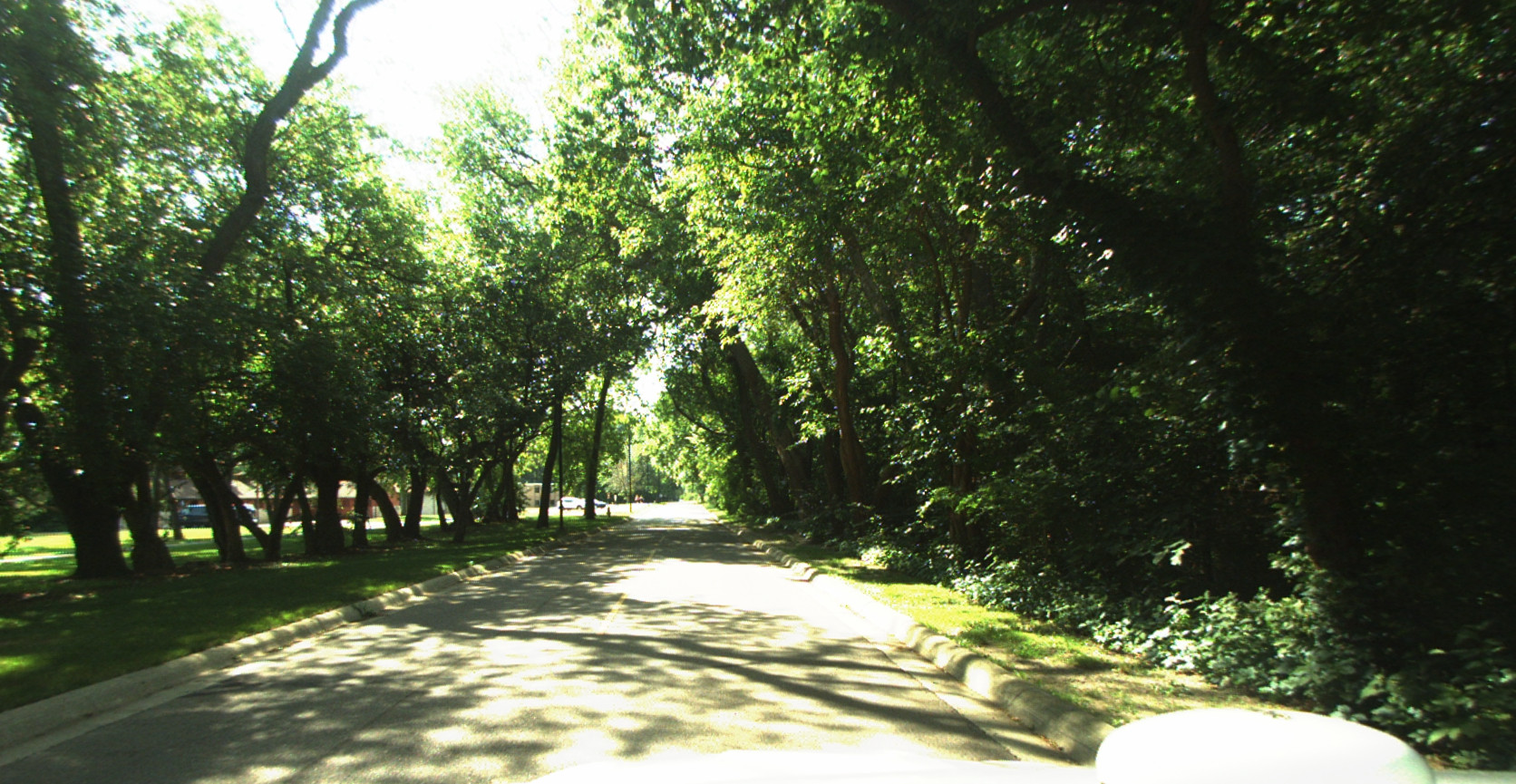}}
	
	\caption{Urban Scenarios}
	\label{fig:urbanscene}
\end{figure}

\section{Scenarios} 

This dataset presents the seasonal variation in weather and lighting conditions experienced throughout the year in urban environments such as the Detroit Metro Area. As shown in Figure \ref{fig:weather_variation}, the same scene can look very different depending on the weather conditions. Any self-driving vehicle platform should be able to operate safely throughout the year. This variation in the dataset can help researchers design better algorithms that are robust to such weather and lightning changes. Besides, this dataset also captures different traffic conditions like construction zones, under-pass, tunnels, airport, residential areas, highways and country side as shown in Figure \ref{fig:urbanscene}. Most importantly, we have used multiple autonomous vehicle platforms collecting this data simultaneously, that will help in opening new research avenues in the area of collaborative autonomous driving. 

\section{Summary}
We present a multi-agent time-synchronized perception (camera/lidar) and navigational (GPS/INS/IMU) data from a fleet of autonomous vehicle platforms travelling through a variety of scenarios over the period of one year. This dataset also includes 3D point cloud and ground reflectivity map of the environment along with ground truth pose of the host vehicle obtained from an offline SLAM algorithm. We also provide ROS based tools for visualization and easy data manipulation for scientific research. We believe that this dataset will be very useful to the research community working on various aspects of autonomous navigation of vehicles. This is first-of-a-kind dataset containing data from multiple vehicles driving through an environment simultaneously, therefore, it will open new research opportunities in collaborative autonomous driving.

\begin{acks}
This data set is the outcome of a joint effort, starting with the foresight and guidance provided by our Senior Technical Leader Dr. Jim McBride and our manager Tony Lockwood. This work was made possible by the diligence and persistence of the Ford / AV LLC team members. Preparing the vehicles, maintaining the code and sensor calibration, and taking a half day on every run to collect the data reflects the values of this team and the desire to make a lasting contribution to the field.  During this time, starting in June 2017 and extending to July 2018 developers were also qualified safety drivers and test engineers, so these data drives represented a significant investment of time for colleagues also delivering on immediate team objectives.  We wish to thank those without whom this dataset could not be made available to the community: Peng-yu Chen, Thaddeus Townsend, Jakob Hoellerbauer, Thomas Iverson, Sharath Nair, Kevin Walker, Matt Warner, Matt Wilmes and Lu Xu. We thank Rob Lupa and his team at Quantum Signal AI for designing the 3D Ford Fusion model released with this dataset. We would also like to thank Daniel Pierce and his communications team at Ford AV LLC for helping out with the public facing website and information articles.
\end{acks}

\bibliographystyle{SageH}
\bibliography{refs}

\end{document}